\documentclass[manuscript]{acmart}
\AtBeginDocument{%
  }

\usepackage{enumitem}

\usepackage[table]{xcolor}
\usepackage{longtable}
\usepackage{multirow}

\newcommand{\andreas}[1]{\textcolor{orange}{Andreas: #1}}

\begin{document}

\title{CoGITAtion: Collaborative Gaze-based Interpretable Task-adaptive Representation for Privacy-sensitive Applications} 

\author{Mayar Elfares}
\correspondingauthor
\email{mayar.elfares@vis.uni-stuttgart.de}
\orcid{1234-5678-9012}
\author{Yasmine Elfares}
\email{Yasmine.Elfares@glasgow.ac.uk}
\author{Andreas Bulling}
\email{andreas.bulling@vis.uni-stuttgart.de}
\affiliation{%
  \institution{University of Stuttgart}
  \country{Germany}
}

\renewcommand{\shortauthors}{Elfares et al.}

\begin{abstract}
Gaze-based analysis provides rich behavioural information for privacy-sensitive applications, but deployment is challenged by heterogeneous, scarce, and distributed data. 
We present CoGITAtion (Collaborative Gaze-based Interpretable Task-Adaptive Representation), a framework for privacy-preserving inference over vertically distributed gaze data. 
CoGITAtion transforms variable-length gaze recordings into compact, interpretable gaze-transition graphs and descriptors computed locally by each data owner. 
Task-adaptive models accommodate heterogeneous and missing feature blocks through task and availability gating, avoiding unnecessary private computation. 
Results show that shared components capture patterns that generalise across tasks, while task-specific components provide capacity without increasing model size. 
Public availability gating further reduces encrypted computation by removing branches determined by publicly known task and owner availability without altering predictions. 
Overall, CoGITAtion enables gaze model reuse, efficient privacy-preserving deployment, and robust inference under incomplete owner availability while retaining interpretable gaze representations.

\end{abstract}

\maketitle

\section{Introduction}\label{sec:intro}

\begin{figure}
    \centering
    \includegraphics[width=0.8\linewidth, height = 5cm]{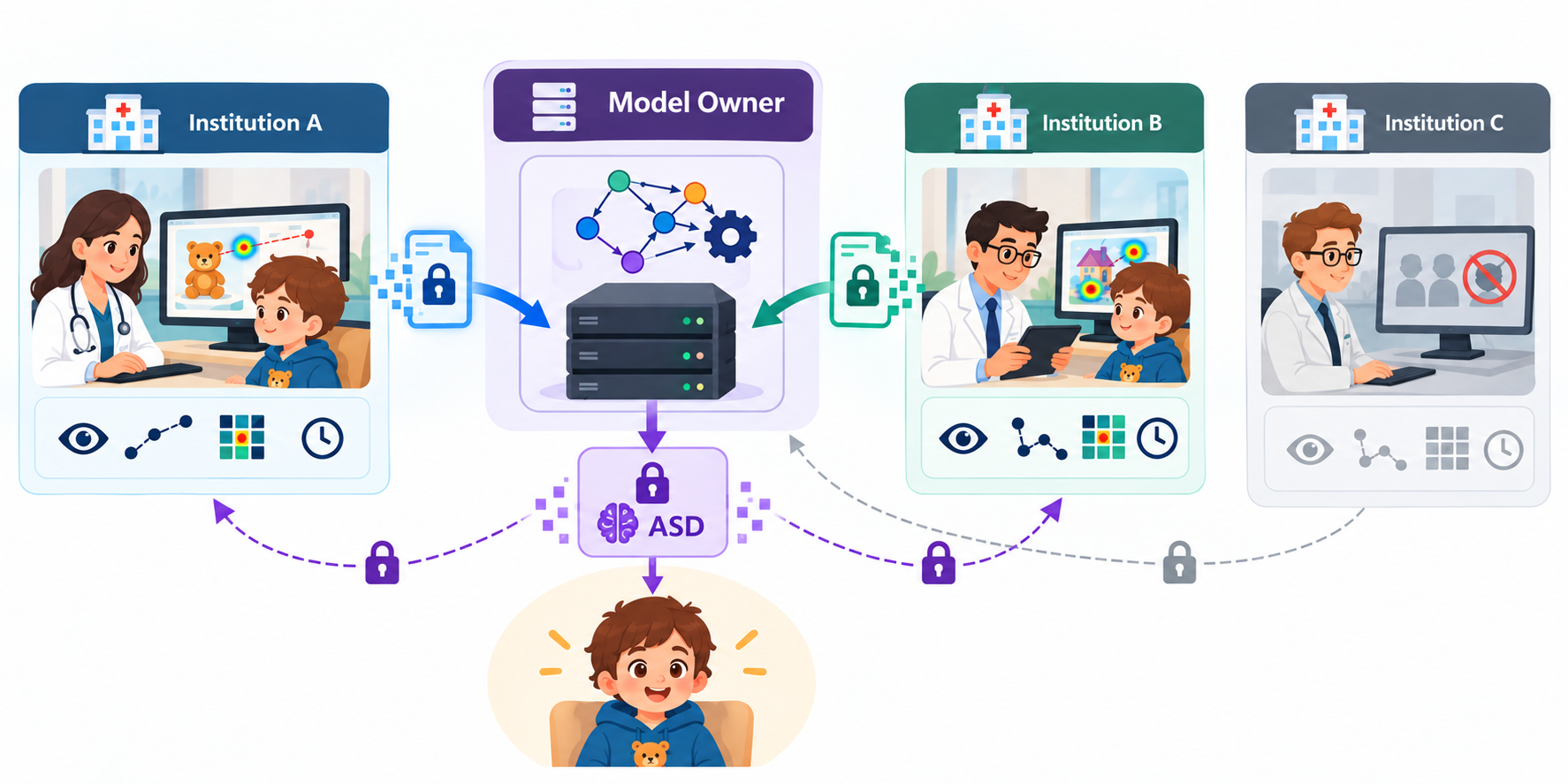}
    \caption{Overview of CoGITAtion. Two or more institutions collect complementary behavioural data from the same child. The data is securely transmitted to the model owner(s), where task-adaptive gaze-graph inference is performed. The private prediction (ASD) is then returned to the authorised parties. Institution C is connected to the framework but does not hold data for the child and therefore contributes no input to the inference process thanks to our gating mechanism.}
    \label{fig:teaser}
\end{figure}

Gaze behaviour provides rich information about human attention, perception, and behaviour \cite{vertegaal2003attentive, bulling2010toward}.  
This information may also be used in sensitive application domains, including the assessment of neurological and neurodevelopmental conditions, mental-health-related analysis, or software development with proprietary and confidential code \cite{elfares2025guidelines, al2026preliminary}. 
Unlike conventional approaches that depend on verbal responses, questionnaires, or structured observations, gaze-based analysis can provide behavioural measurements directly from how individuals visually interact with stimuli \cite{iwauchi2023eye}.

Despite its potential, the adoption of gaze-based analysis in sensitive applications remains limited. 
One fundamental challenge is the heterogeneity, scarcity, and distributed ownership of gaze data \cite{elfares23federated, holmqvist2012quality}. There is no universally established representation or acquisition protocol, and gaze recordings can differ substantially across eye trackers, experimental protocols, stimuli, interfaces, and behavioural tasks. 
Even within the same application domain, the same condition may be assessed using different paradigms, while programming-related gaze behaviour may depend on the particular code, interface, or development environment being inspected \cite{elfares2025guidelines, al2021emip}. 
These discrepancies hinder the integration of heterogeneous datasets and limit the effectiveness of models that assume a homogeneous input space. 
Moreover, gaze data are often scarce and distributed across multiple data owners, such as hospitals, research institutions, and other organisations, which may hold complementary feature sets for the same individuals, a.k.a. vertically distributed data. 
For instance, different owners may possess gaze statistics, pupil measurements, or task-specific behavioural information. 
Privacy regulations, consent constraints, and institutional data-sharing policies further restrict the centralisation and aggregation of such data for model training \cite{gdpr, eu_ai_act_2024}. 

A second challenge is interpretability.
Many high-performing gaze-analysis approaches learn directly from images, raw scanpaths, or other high-dimensional representations. 
Although such models can capture complex behavioural patterns, their predictions may be difficult to relate to observable gaze behaviour. 
This is particularly problematic in sensitive or regulated settings, where understanding why a model produces a particular prediction can be as important as predictive performance itself \cite{kumar2020challenges}. 
A representation based on interpretable behavioural descriptors can instead provide a direct connection between model inputs and measurable properties of visual exploration, such as revisits, regressions, dwell concentration, transition structure, and reading behaviour.   

Collaborative privacy-preserving computation offers a potential solution to the tension between distributed information and data confidentiality. 
However, it introduces substantial computational and communication overhead. 
Efficient private inference therefore requires not only an appropriate predictive model but also a representation and computational structure tailored to encrypted execution.

To address these challenges, we present \textit{CoGITAtion} -- \underline{Co}llaborative \underline{G}aze-based \underline{I}nterpretable \underline{T}ask-\underline{A}daptive
Representa\underline{tion}. 
The core idea of this novel gaze representation is to enable moving computationally expensive, data-dependent gaze processing to the data owners, and provide the privacy-preserving model with a compact fixed-dimensional representation. 
Each gaze recording is locally transformed into a directed, weighted gaze-transition graph describing transitions between spatial or task-specific regions. 
From this graph, CoGITAtion extracts descriptors capturing complementary aspects of visual exploration, including spatial coverage and density, self-loops and revisits, hubs and attractors, reciprocal and cyclic transitions, transition and dwell entropy, and text-reading behaviour. 
The resulting owner block is then used for private inference rather than the original variable-length gaze sequence.

CoGITAtion further addresses heterogeneity through task-adaptive computation, reusing task-independent gaze-based modules while employing gating mechanisms to selectively activate task-specific modules.
Overall, CoGITAtion provides an important deployment advantage; it enables a unified model to operate across multiple tasks and heterogeneous owner configurations, reducing the need for separate models, compilations, and cryptographic configurations.

We evaluate CoGITAtion across heterogeneous privacy-sensitive applications: neurodevelopmental, medical, and programming datasets \cite{duan2019dataset, cilia2022dataset, song2025emsdataset, wainstein2017adhddataset, rojaslibano2019adhddataset, bednarik2020emip, yasmine2025gazecopilot}. 
Our results show that explicitly separating task-independent and task-specific structure allows shared components to capture reusable patterns across tasks, while task-specific components provide additional capacity only where needed.    
Gating further improves deployment efficiency.
Because task identity and owner availability are known before private inference, corresponding branches can be removed at compile time without changing the model output. 
This provides a lossless reduction in private computation when task or owner availability is heterogeneous.     
Thus, CoGITAtion allows model reuse, simpler deployment, and robustness to missing data owners\footnote{Note that privacy-preserving computation does not compensate for poor-quality data. 
Since CoGITAtion is a statistical learning framework, acquisition artefacts, weak labels, missing data, gaze drift, and other measurement errors can directly affect model performance (Section \ref{sec:experiments}). 
We therefore recommend performing data-quality assessment before applying privacy-preserving computation.}.

\paragraph{Research questions and contributions}

\begin{itemize}[leftmargin = *]
    \item \textbf{Can compact gaze-transition descriptors preserve useful predictive information while reducing the cost of private inference?} We introduce a compact gaze-transition representation that transforms variable-length gaze recordings into fixed-dimensional, interpretable descriptors computed locally by each data owner. By moving graph construction and descriptor extraction outside the encrypted computation, the representation reduces the complexity of private inference while retaining useful predictive information across heterogeneous gaze-analysis tasks.

    \item \textbf{Can shared and task-specific components support heterogeneous tasks without requiring a separate deployment for every task?} 
    We develop two complementary task-adaptive architectures, CoGITAtion-NN and CoGITAtion-Tree, that combine shared representations with task-specific prediction components. Their routing and feature-selection mechanisms accommodate heterogeneous and missing feature blocks while enabling model reuse across multiple tasks and reducing unnecessary computation.

    \item \textbf{Can gating reduce private computation while preserving model outputs and accommodating missing owners?}
    We integrate CoGITAtion with fully homomorphic encryption and secure multi-party computation, introducing task- and availability-aware gating that removes irrelevant computation before private inference. This enables collaborative prediction over heterogeneous feature blocks without requiring data owners to disclose their raw gaze measurements, subject to the security assumptions of the respective protocols.

\end{itemize}

\section{Foundations and Related Work}\label{sec:related_work}
\subsection{Privacy-Preserving Eye Tracking}

Numerous studies in ET research and cognitive science have revealed that human eye movements can provide insights into a user's mental state \cite{bulling2011recognition, bulling2014cognition}, levels of text comprehension \cite{ahn2020towards}, intention \cite{lethaus2013comparison, sattar2020deep, zhang2022predicting},  mind-wandering tendencies \cite{huang19_mm,zermiani22_edueye}, recallability \cite{wang2024visrecall++}, bottom-up \cite{borkin2015beyond, shin2022scanner} as well as top-down attention \cite{gomez2016fauxvea, lalle2016predicting, polatsek2018exploring} from their eye movements.
The richness and granularity of information that can be inferred from gaze data have raised significant privacy concerns.

A large body of prior work addresses gaze privacy through system-level and behavioural interventions evaluated empirically rather than through formal guarantees. 
Research has focused on secure sensing pipelines, privacy-preserving rendering, and protected interaction protocols for ET systems \cite{li2021kaleido, david2021privacy, david2021towards, david2022your},  the raw sensor data itself \cite{chaudhary2020privacy}, the use of synthetic data \cite{bozkir2020privacy}, and multi-modal adversaries \cite{aziz2025exploring}.
In contrast to empirical approaches, formal methods in this space primarily adopt differential privacy (DP) as a framework \cite{dwork2014algorithmic}, offering mathematically bounded guarantees on the information that any computation over gaze signals can reveal about individual users \cite{steil2019privacy, bozkir2021diffrential,ren2024dpgazesynth, david2021privacy, david2022your, elfares26gaze3p}. 
A few others adopted secure multi-party computation and homomorphic encyption-based defences \cite{elfares2024privateyes, elfares25qualiteye, ozdel2024privacy}.

\subsection{Privacy-Preserving Mechanisms}

We therefore focus on two widely adopted privacy-preserving computation paradigms: MPC and FHE.
For FHE, we specifically consider TFHE due to its efficient bootstrapping and support for Boolean circuits, making it well suited for privacy-preserving inference involving nonlinear operations and comparisons.

Multi-party computation (MPC) enables multiple parties to jointly compute a function over their private inputs while revealing no information beyond the output, subject to the protocol's security assumptions. Common approaches include Yao's garbled circuits \cite{yao1982protocols} and secret sharing \cite{damgaard2012spdz}. MPC provides privacy and correctness but typically incurs communication and coordination overhead.

Similarly, fully homomorphic encryption (FHE) enables arbitrary computations on encrypted data without decrypting it. Given a plaintext ($m$) and its ciphertext ($c=\mathsf{Enc}(m)$), homomorphic evaluation satisfies ($\mathsf{Dec}(\mathsf{Eval}(f,c))=f(m)$). Unlike partially homomorphic schemes, FHE supports arbitrary circuits combining addition and multiplication.  Prominent schemes include BFV and BGV for exact arithmetic, CKKS for approximate arithmetic, and TFHE, which supports efficient Boolean and integer operations through programmable bootstrapping \cite{gentry2009fully}.
Bootstrapping refreshes ciphertexts to control noise growth, enabling computations of arbitrary depth, but introduces substantial computational overhead. FHE therefore enables non-interactive computation by untrusted servers while preserving data confidentiality, at the cost of increased computational complexity.

\section{Problem Formulation}\label{sec:problem}
\textbf{Entities.} A set of data owners $\mathcal{O}=\{O_1,\dots,O_M\}$ hold measurements of the same people, i.e. vertically distributed data. 
Owner $O_m$ holds feature blocks $\mathcal{B}_m$ (e.g., raw-gaze features, pupil statistics, AOI statistics or task behaviour), and $\mathbf{x}_{i,b}\in\mathbb{R}^{d_b}$ denotes block $b$ of person $i$. 
A set of assessment tasks $\mathcal{T}$ (eye-tracking paradigms and conditions, e.g. searching or browsing) is supported. 
Task $t$ uses the blocks $\mathcal{B}_t\subseteq\bigcup_m\mathcal{B}_m$. 
A \emph{model owner} trains and serves a model $f_\theta$. 

\vspace{0.25cm}
\noindent\textbf{Inference under MPC.} For person $i$ and administered task $t$, each owner secret-shares its blocks
$[\mathbf{x}_{i,b}]=(\mathbf{x}_{i,b}^{(1)},\dots,\mathbf{x}_{i,b}^{(n)})$, $b\in\mathcal{B}_t\cap\mathcal{B}_m$, among $n$
computing parties $P_1,\dots,P_n$, such that $\sum_{j}\mathbf{x}_{i,b}^{(j)}=\mathbf{x}_{i,b}$ and every share is
authenticated by an information-theoretic MAC; the model owner likewise secret-shares the parameters $[\theta]$, so the
model is protected as well.
The parties jointly compute $[\hat p_{i,t}]=f_{[\theta]}(t,\{[\mathbf{x}_{i,b}]\})$ on shares, using preprocessed Beaver triples for multiplications and secure comparisons, without any party learning an input, a parameter or the result; after a MAC check, only the result recipient (e.g. the clinic requesting the assessment) reconstructs $\hat p_{i,t}$ from the parties' output shares.

\vspace{0.25cm}
\noindent\textbf{Inference under FHE.} For person $i$ and administered task $t$, each owner encrypts its blocks $\llbracket\mathbf{x}_{i,b}\rrbracket$, $b\in\mathcal{B}_t\cap\mathcal{B}_m$.
The server computes $\llbracket\hat p_{i,t}\rrbracket=f_\theta(t,\{\llbracket\mathbf{x}_{i,b}\rrbracket\})$ without decrypting, and only the key holder (e.g. the clinic requesting the assessment) decrypts $\hat p_{i,t}$.
Under TFHE, additions and multiplications by plaintext constants are cheap, while each non-linear activation on a quantised value requires one programmable bootstrap. 
We therefore measure encrypted cost by (i) the number of PBS per inference, which dominates latency, (ii) the number of encrypted inputs, which drives ciphertext volume and client-side work, and (iii) the number of active parameters, a proxy for circuit size and memory.

\noindent\textbf{Objective.} Maximise discrimination for every task while minimising per-task encrypted cost, accommodating tasks with different and possibly missing blocks, and without any owner disclosing raw measurements.

\section{CoGITAtion}\label{sec:method}

\subsection{Gaze Graphs Descriptors}
The local gaze graph descriptors constitutes the first processing stage of our method as well as our main contribution. 
This stage is performed locally by each data owner in plaintext, prior to any encryption using fully homomorphic encryption (FHE) or secret sharing for multi-party computation (MPC).

Each data owner possesses different eye-tracking recordings.
Because individual recordings may contain thousands to millions of samples and may vary substantially in duration, directly processing the raw sequences within a privacy-preserving computation would be computationally expensive.
Therefore, each recording is transformed locally into \textbf{a fixed-dimensional numerical representation, referred to as an owner block}. 
Only these derived descriptors are subsequently provided to the private computation. 

\begin{figure}
    \centering
    \includegraphics[width=1\linewidth]{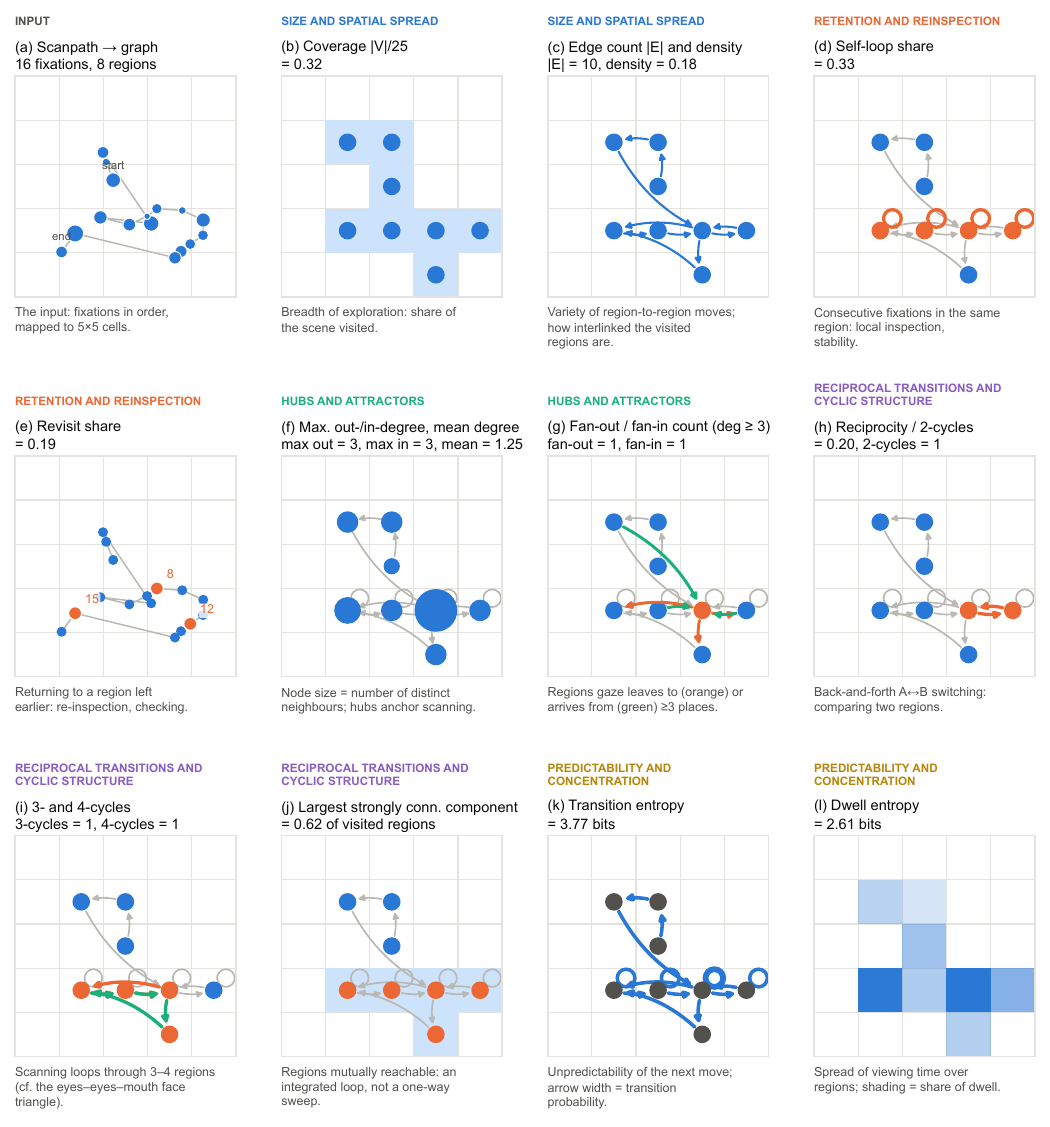}
    \caption{Graph-based gaze representation and extracted structural descriptors. 
    }
    \label{fig:descriptors}
\end{figure}

For datasets with pre-segmented fixations, these are used directly, otherwise, fixations are identified from raw gaze recordings using the velocity-threshold (I-VT) method ($\geq$60 ms).  
Gaze coordinates are normalised to ([0,1]), yielding a scanpath
$s=(v_1,\ldots,v_n),$
where each ($v_k$) denotes a fixation position and duration.
The scanpath is subsequently represented as a directed, weighted graph describing transitions between spatial regions of the visual field. We handle diffrent resolutions in Section \ref{sec:experiments}.
We propose two descriptors for image- and text-based stimuli, when available:
\begin{itemize}[leftmargin = *]
    \item \textbf{Nodes:}
    For images, the visual field is partitioned into a \(5\times5\) regular grid comprising 25 equally sized cells\footnote{Equal cells give every dataset the same nodes despite very different stimuli; 25 cells keep the graphs of short scanpaths (tens of fixations) dense enough for degree, reciprocity, cycle and entropy descriptors to be informative, and an odd grid places one cell at the screen centre, where gaze is concentrated (in addition to typical central fixation during calibration). 
    The grid size was fixed a priori and not tuned.}. 
    Each fixation \(v_k\) is assigned to a grid cell \(c(v_k)\). 
    Every cell visited at least once constitutes a graph node.
    For text, a spatial \(5\times5\) grid is less informative because the relevant unit of analysis is the text line rather than an arbitrary screen region. 
    Therefore, fixations are additionally mapped to text lines (the nearest line according to its vertical centre), which serve as task-specific areas of interest.

    \item \textbf{Edges:}
    Directed edges represent transitions between consecutive fixations. 
    The weight of the edge from cell \(j\) to cell \(k\) is defined as $W_{jk}= \#\{\text{transitions from cell }j\text{ to cell }k\}.$
    Thus, \(W_{jk}\) corresponds to the number of times a fixation located in cell \(j\) is immediately followed by a fixation located in cell \(k\). 
    When \(j=k\), the transition constitutes a \textbf{self-loop}, representing repeated fixation within the same spatial region. 
    For each cell \(j\), the \textbf{dwell time} \(D_j\) is defined as the cumulative duration of all fixations assigned to that cell (Figure \ref{fig:descriptors}).
\end{itemize}

Consequently, the resulting representation is a directed, weighted graph.
Its topology captures characteristics of visual exploration, including the spatial concentration of gaze, the degree of systematicity, and the frequency with which previously inspected regions are revisited (see Figure \ref{fig:descriptors}).
The owner further computes 20 conventional gaze statistics, including fixation counts and duration statistics, saccade amplitudes, scanpath length, the mean and spatial dispersion of gaze positions, distance from the screen centre, and bounding-box area.
We additionally propose the following descriptors:

\newpage
{\Small
\setlength{\tabcolsep}{4pt}        
\renewcommand{\arraystretch}{0.9}  
\setlength{\LTpre}{0pt}            
\setlength{\LTpost}{0pt}           

\begin{longtable}{p{0.25\textwidth} p{0.23\textwidth} p{0.45\textwidth}}
\textbf{Category} & \textbf{Measure} & \textbf{Description} \\
\hline
\endfirsthead

\textbf{Category} & \textbf{Measure} & \textbf{Description} \\
\hline
\endhead

\multirow[t]{3}{=}{Size and Spatial Spread}
& Coverage
& Proportion of the grid cells that were visited. \\
& Edge count
& Number of distinct transitions between different cells. \\
& Density
& Ratio between the number of observed directed transitions and the number of possible directed transitions among the visited cells. Higher density indicates transitions between a larger proportion of possible cell pairs. \\
\hline

\multirow[t]{2}{=}{Retention and Reinspection}
& Self-loop share
& Proportion of transitions that remain within the same cell. Higher values indicate repeated fixation or reinspection within a spatial region. \\
& Revisit share
& Proportion of fixations that return to a previously visited cell, excluding transitions that merely remain within the same cell. \\
\hline

\multirow[t]{5}{=}{Hubs and Attractors}
& Maximum out-degree
& Maximum number of distinct destination cells reached from a single cell. \\
& Maximum in-degree
& Maximum number of distinct origin cells from which a given cell is reached. \\
& Mean degree
& Mean number of distinct destinations per cell. \\
& Fan-out count
& Number of cells having at least three distinct outgoing destinations. \\
& Fan-in count
& Number of cells having at least three distinct incoming origins. \\
\hline

\multirow[t]{5}{=}{Reciprocal Transitions and Cyclic Structure}
& Reciprocity
& Proportion of transitions for which the corresponding reverse transition is also observed. High reciprocity indicates frequent alternation between pairs of regions. \\
& 2-cycles
& Number of reciprocal two-node transition pairs. \\
& 3-cycles
& Number of closed three-node cycles of the form \(j\rightarrow k\rightarrow l\rightarrow j\). \\
& 4-cycles
& Number of closed four-node cycles. \\
& Largest strongly connected component
& Proportion of visited cells belonging to the largest strongly connected component, i.e., the largest set of cells that are mutually reachable through observed transitions. The cycle-based measures provide information about repeated and structured scanning patterns, whereas the size of the largest strongly connected component reflects the degree to which visual exploration forms a mutually connected network. \\
\hline

\multirow[t]{2}{=}{Predictability and Concentration}
& Transition entropy
& Shannon entropy, with logarithm base 2, calculated from the transition-frequency distribution. Lower values indicate a smaller number of repeatedly used transitions, whereas higher values indicate a larger and more evenly distributed set of transitions. \\
& Dwell entropy
& Shannon entropy of the dwell-time distribution across visited cells. Lower values indicate concentration of viewing time in a small number of regions, whereas higher values indicate a more uniform distribution of viewing time. \\
\hline

\multirow[t]{12}{=}{Text Reading}
& On-text share
& Proportion of fixations occurring on text lines. \\
& Line coverage
& Proportion of text lines fixated at least once. \\
& Next-line rate
& Proportion of line transitions directed to the immediately subsequent line. \\
& Same-line rate
& Proportion of transitions that remain on the same line. \\
& Regression rate
& Proportion of transitions returning to a previously viewed line. \\
& Mean line jump
& Mean number of lines skipped during a transition between different lines. \\
& Maximum line jump
& Maximum observed distance between consecutively fixated lines. \\
& Line-dwell entropy
& Normalised Shannon entropy of the dwell-time distribution across text lines. \\
& Revisit share
& Proportion of line visits corresponding to previously inspected lines. \\
& Dwell concentration
& Proportion of total reading time allocated to the single most frequently viewed line. \\
& Vertical position
& Mean normalised vertical position of the fixated lines, where 0 corresponds to the top and 1 to the bottom of the text. \\
& Lines per second
& Number of line changes per second of fixation time. \\
\hline

\end{longtable}
}


 \newpage
Each data owner contributes one or more feature blocks depending on the data available to that owner. 
Different institutions may possess different modalities, such as gaze recordings, pupil measurements, or text-aligned gaze logs. 
During our multimodal aggregation, each block is treated as an independent contribution. 

This representation provides three key benefits: (i) \textbf{interpretability}, as semantically meaningful descriptors directly link model predictions to observable gaze behaviour (e.g., high regression and low next-line rates indicate non-linear text reading) (Section \ref{sec:experiments}); 
(ii) \textbf{data minimisation}, as each owner contributes only a fixed-dimensional feature vector, while raw and variable-length gaze sequences remain local; 
and (iii) \textbf{computational efficiency} in privacy-preserving settings, as graph construction is performed locally, avoiding costly data-dependent comparisons and lookups within FHE/MPC circuits, where such operations are computationally expensive\footnote{Note that, the descriptor-based representation involves an inherent information-compression trade-off. Transforming a complete scanpath into a fixed-dimensional summary inevitably discards some temporal information. 
We show later in our results that this loss did not adversely affect the tasks.}.

\subsection{Task-Adaptive Computation}
Computational efficiency can be further improved by (1) selectively omitting features that are unavailable or deemed non-informative for a given task, and (2) deploying a single unified model to handle all tasks, thereby reducing redundant model parameters and computational overhead. After gaze descriptors are extracted, they are fed into a task-specific computation graph that integrates the available heterogeneous feature blocks within a shared computational framework while adapting the processing pathway to the requirements of each task.
We evaluate two model families commonly employed in privacy-preserving machine learning: neural networks (NNs)--CoGITAtion-NN-- and decision trees--

\textbf{CoGITAtion-NN.} Each feature block \(b\) is processed by a block-specific encoder, producing an embedding \(e_b\). 
For task \(t\), a route \(R_t\) determines the active blocks and applies feature masks \(m_{t,b}\). 
The resulting embeddings are summed with a task-specific bias and passed through a shared trunk and task-specific prediction head:
$e_b =
W_b^{(2)}
\sigma\left(
W_b^{(1)}(m_{t,b}\odot x_b)+c_b^{(1)}
\right)+c_b^{(2)}$,
$\hat p_t =
\operatorname{sig}\left(
g_t\left[
\sigma\left(
T\sigma\left(\beta_t+\sum_{b\in R_t}e_b\right)+c_T
\right)
\right]
\right)$.
Here, \(\sigma\) denotes ReLU, with hidden dimensions \(H=24\), embedding dimension \(E=16\), and task-head dimension \(H_h=8\) in accordance with the input size. 
Missing blocks contribute zero through the route definition. 
For efficiency, during task adaptation, blocks with low drop-importance (\(\leq\tau=0.002\)) are removed, and features outside the set accounting for \(\rho=0.9\) of permutation importance are masked (Section \ref{sec:experiments}). 
Shared parameters are frozen while task-specific parameters are adapted and subsequently merged into the unified model.

\textbf{CoGITAtion-Tree.} CoGITAtion-Tree uses a shared ensemble together with task-specific trees. 
All tasks share a common input representation containing the feature blocks, availability indicators, and task indicator.  
The prediction function is
$F_t(u)=
\sum_{s=1}^{S}h_s(u)
+
\sum_{k=1}^{K}h_k^t(u),
\qquad
\hat p_t=\operatorname{sig}(F_t(u))$,
with (S=K=50) trees of depth 3. 
Task-specific trees are boosted from the shared ensemble. 
Since task identity, availability indicators, and missing-block indicators are known when the task is administered, splits on these variables act as \textbf{public gates}. 
These gates can therefore be folded before inference, retaining only the branches relevant to the administered task while preserving the original prediction exactly.

\subsection{Privacy-Preserving Back-Ends}
With an efficient representation and task-specific computational models established, we next integrate them with privacy-preserving computation, using both FHE and MPC. 
Our main contribution at this stage is not the cryptographic primitives themselves, but the \textbf{task-specific gating of the computation}: the model selectively activates the heterogeneous feature blocks available from different data owners, allowing each task to operate on its relevant subset of potentially scarce and incomplete inputs while avoiding unnecessary private computation.

\textbf{In the FHE setting}, for CoGITAtion-NN, a task route \(R_t\) specifies the active blocks and feature masks.  
Since each ReLU on an encrypted value requires one programmable bootstrapping (PBS), a route containing \(|R_t|\) active blocks requires
$|R_t|(H+E)+2E+H_h+1$
PBS operations, whereas the task-private mode, which evaluates all block encoders and task heads, requires 
$|B|(H+E)+|T|(2E+H_h+1)$.
Thus, task-specific gating directly reduces the encrypted computation when only a subset of heterogeneous blocks is available or relevant.
For CoGITAtion-Tree, tree ensembles are evaluated in tensor form. 
Each internal node requires an encrypted comparison and each leaf a lookup; consequently, a tree with \(n\) internal nodes requires approximately \(2n+1\) PBS operations. 

\textbf{In the MPC setting}, the participating parties hold additive shares \([x]=(x_1,\ldots,x_n)\), with \(\sum_i x_i=x\) in \(\mathbb{F}_p\), together with authenticated MAC shares under a shared global key. 
Data owners secret-share their feature blocks, while the model owner can secret-share thresholds, leaf values, and network weights, thereby protecting both data and model parameters. 
Additions are local; secure multiplications consume Beaver triples, and opened values are authenticated through batched MAC checks. 
Triples and random bits are generated in an input-independent offline phase.
For CoGITAtion-NN, linear layers with secret weights require secure multiplications, while each ReLU requires a secure comparison and multiplication; fixed-point arithmetic additionally requires truncation. 
In contrast to FHE, where plaintext weights make linear layers comparatively inexpensive, feature-level gating is particularly beneficial in MPC because eliminating unavailable or irrelevant features directly reduces the number of secure multiplications.
For CoGITAtion-Tree, features and thresholds are quantised to 8 bits per feature. 
Each internal node evaluates a secure comparison \([b]=[x_j<\theta]\) using bit decomposition. 
Comparisons across trees are batched, while path indicators are propagated level by level,
$[\iota_{\mathrm{yes}}]=[\iota][b],\qquad
[\iota_{\mathrm{no}}]=[\iota]-[\iota][b]$,
requiring one multiplication per node. 
The final margin is obtained by summing the leaf values weighted by their path indicators. Consequently, the number of interaction rounds depends primarily on the tree depth and comparison depth rather than on the number of trees. 
As in FHE, public task and availability indicators allow irrelevant branches to be folded away before private evaluation.





\section{Experiments}\label{sec:experiments}

\subsection{Datasets}
We evaluate the proposed approach on a heterogeneous collection of privacy-sensitive eye-tracking datasets:
The medical datasets comprise image-, video-, and web-based viewing tasks involving clinical populations, while the programming datasets contain gaze recordings collected during code comprehension tasks, including proprietary or controlled-access development environments. 
This diversity allows us to evaluate the framework under heterogeneous modalities, task definitions, and data-owner configurations.

The medical datasets include Autism Spectrum Disorder (ASD) classification (Duan et al \cite{duan2019dataset} and Cilia et al. \cite{cilia2022dataset}), schizophrenia classification (Song et al.'s EMS \cite{song2025emsdataset}), and ADHD classification of participants off or on medication (Wainstein et al. \cite{wainstein2017adhddataset} and Rojaslibano et al. \cite{rojaslibano2019adhddataset}). 
The programming datasets comprise EMIP \cite{bednarik2020emip} and GazeCopilot \cite{yasmine2025gazecopilot} for code readability, programming-expertise, and comprehension tasks. 

Note that, in all our results, EMIP results did not follow the other datasets' patterns.
We believe that EMIP is especially challenging primarily due to weak labels (expertise is self-reported, while comprehension is based on a single multiple-choice question, introducing substantial noise.), limited information per sample (Short 15-second windows and small programs capture only part of participants’ overall code-reading behaviour.), and data-quality issues (Multi-site mobile eye tracking and uncorrected gaze drift may introduce acquisition-related noise and line-assignment errors.).
We hence recommend using data quality measures \cite{elfares25qualiteye, holmqvist2012quality} before running any privacy-preserving mechanism such as ours.

\subsection{Results}
We evaluate the proposed approach under plaintext, FHE, and MPC settings. 
We first compare our representation against learning-based baselines to assess predictive performance and computational efficiency. 
We then conduct ablation studies to quantify the contribution of individual components and evaluate interpretability.

\subsubsection{\textbf{Our respresentation vs. learned representations}}\label{sec:ev-gin}

We constructed our gaze graphs using explicitly defined features to enhance interpretability; alternatively, these features could be learned directly from the data.
As that baseline we use a graph isomorphism network (GIN)~\cite{xu2018gin}\footnote{We specifically used GIN over other GNNs because it is the most expressive standard message-passing GNN and its operations are the cheapest to run under encryption \cite{effendi2024privacy}. 
Although neural networks and other learning paradigms can be deployed end-to-end, their execution under MPC or FHE incurs substantial computational overhead. Splitting the architecture, as in split learning, may reduce this cost but introduces additional privacy risks from intermediate representations and communication overhead.} 
with different variants to match our gaze graphs:(i) \textbf{GIN-Set}: the 32 fixation nodes without edges; each fixation is encoded on its own and the node embeddings are pooled, (ii) \textbf{GIN-Chain} (main variant): a temporal chain in which each fixation is linked to the next one in time. 
Its adjacency is identical for every person, hence public, so aggregation is a product of a plaintext matrix with
encrypted (or secret-shared) node features, (iii) \textbf{GIN-Masked}: GIN-Chain with padding nodes of short scanpaths excluded exactly by a mask; the mask depends on the person's number of fixations and must therefore be encrypted, (iv) \textbf{GIN-Grid}: 25 nodes, one per cell of the $5{\times}5$ screen grid, with edge weights given by the person's gaze transitions between cells; the adjacency is derived from the person's own data and must be encrypted, so every aggregation becomes a ciphertext--ciphertext product.
GIN-Set and GIN-Chain are the only constructions whose adjacency is public and therefore cheap under encryption. 

\begin{table*}[t]
\centering
\caption{Sample AUC and private-inference cost for models operating
on engineered gaze graphs and descriptors (CoGITAtion) and learned
features (GINs). FHE (TFHE): encrypted inputs per inference, programmable
bootstraps (PBS), PBS depth (sequential bootstrapping layers, which bound
latency when all PBS of a layer run in parallel) and ciphertext--ciphertext
products (each needs extra PBS; plaintext-weight products need none).
MPC uses SPDZ with 3 parties and WAN settings of 50\,ms/100\,Mb/s.
Ranges are over the 11 tasks; CoGITAtion-NN inputs are the pruned route's
features, averaged over folds.}
\label{tab:ev-gin-acc}
\setlength{\tabcolsep}{4pt}
\small
\begin{tabular}{@{}lcccccccc@{}}
\toprule
& \multicolumn{4}{c}{\textbf{Gaze graphs and descriptors}}
& \multicolumn{4}{c}{\textbf{Learned features}} \\
\cmidrule(lr){2-5}\cmidrule(lr){6-9}
& \multicolumn{2}{c}{\textbf{CoGITAtion}}
& \multicolumn{2}{c}{\textbf{Per-task}}
& \multicolumn{4}{c}{\textbf{GIN}} \\
\cmidrule(lr){2-3}\cmidrule(lr){4-5}\cmidrule(lr){6-9}
\textbf{Task / Metric} & \textbf{Tree} & \textbf{NN} & \textbf{Tree} & \textbf{NN} & \textbf{Set} & \textbf{Chain} & \textbf{Masked} & \textbf{Grid} \\
\midrule
Duan (ASD)                & 0.740 & 0.712 & 0.742 & 0.736 & 0.717 & 0.723 & 0.678 & 0.637 \\
Cilia (ASD)                 & 0.724 & 0.600 & 0.728 & 0.677 & 0.693 & 0.700 & 0.681 & 0.594 \\
EMS (schizophrenia)         & 0.809 & 0.780 & 0.807 & 0.819 & 0.675 & 0.670 & 0.665 & 0.640 \\
ADHD off-medication         & 0.756 & 0.627 & 0.756 & 0.623 & 0.645 & 0.680 & 0.691 & 0.548 \\
ADHD on-medication          & 0.739 & 0.537 & 0.741 & 0.536 & 0.715 & 0.732 & 0.735 & 0.555 \\
\midrule
EMIP expertise (rectangle)  & 0.569 & 0.533 & 0.564 & 0.528 & 0.590 & 0.598 & 0.592 & 0.553 \\
EMIP expertise (vehicle)    & 0.575 & 0.504 & 0.577 & 0.579 & 0.563 & 0.558 & 0.564 & 0.540 \\
EMIP correct (rectangle)    & 0.462 & 0.479 & 0.474 & 0.502 & 0.591 & 0.589 & 0.585 & 0.559 \\
EMIP correct (vehicle)      & 0.441 & 0.472 & 0.448 & 0.486 & 0.483 & 0.485 & 0.487 & 0.509 \\
GazeCopilot readability     & 0.954 & 0.948 & 0.952 & 0.952 & 0.808 & 0.784 & 0.775 & 0.753 \\
GazeCopilot comprehension   & 0.688 & 0.647 & 0.693 & 0.653 & 0.678 & 0.668 & 0.675 & 0.570 \\
\midrule
Mean, medical (5)           & 0.754 & 0.651 & 0.755 & 0.678 & 0.689 & 0.701 & 0.690 & 0.595 \\
Mean, programming (6)       & 0.615 & 0.597 & 0.618 & 0.617 & 0.619 & 0.614 & 0.613 & 0.581 \\
\midrule
\multicolumn{9}{l}{\textit{Private-inference cost}} \\
\midrule
FHE encrypted inputs        & 37--59 & 6--16 & 37--59 & 37--59 & 128 & 128 & 128 & 700 \\
FHE PBS                     & 1,348--1,467 & 84--121 & 638--743 & 121--201 & 4,113 & 4,113 & 6,161 & 43,217 \\
FHE PBS depth               & 2 & 6 & 2 & 6 & 6 & 6 & 8 & 8 \\
FHE ct$\times$ct products   & 0 & 0 & 0 & 0 & 0 & 0 & 2,048 & 40,000 \\
MPC comparisons             & 662 & 76 & 326 & 125 & 4,113 & 4,113 & 4,113 & 3,217 \\
MPC multiplications         & 21,168 & 13,114 & 10,429 & 22,516 & 744,556 & 744,556 & 746,604 & 622,380 \\
MPC rounds                  & 15 & 41 & 15 & 41 & 57 & 57 & 59 & 59 \\
MPC communication (kB/party)& 1,390 & 825 & 685 & 1,417 & 46,667 & 46,667 & 46,795 & 39,003 \\
MPC WAN latency             & 0.86\,s & 2.12\,s & 0.81\,s & 2.17\,s & 6.67\,s & 6.67\,s & 6.78\,s & 6.15\,s \\
\bottomrule
\end{tabular}
\end{table*}

Across all medical tasks, CoGITAtion-Tree outperforms all four GIN variants. 
CoGITAtion-NN is less accurate, falling below the best GIN on four tasks. 
On GazeCopilot, both CoGITAtion models outperform every GIN. 
EMIP code reading is the exception: GINs perform best on programming tasks, suggesting that fixation order captures information lost by 15,s summary descriptors \cite{busjahn2015eye}. 
Among GIN variants, temporal edges help primarily for ADHD, while padding masks rarely improve performance and data-dependent grid edges generally perform worst.

Because gaze graphs are constructed before encryption, CoGITAtion requires only 831--1{,}467 PBS for CoGITAtion-Tree and 84--121 for CoGITAtion-NN, compared with 4{,}113 for GIN-Set/Chain, 6{,}161 with encrypted padding masks, and 43{,}217 with encrypted grid adjacency. Thus, CoGITAtion is approximately $3$--$40\times$ cheaper under FHE.

Under MPC, GIN requires approximately 4{,}100 comparisons, over 740{,}000 multiplications, 46,MB of online communication per party, and an estimated 6.7,s over a WAN. 
CoGITAtion-Tree requires 603 comparisons, 19{,}294 multiplications, 1.3,MB, and 0.85,s; CoGITAtion-NN takes 2.1,s. 
Although grid adjacency is less costly under MPC than FHE, it remains the least accurate variant.

Overall, \textbf{summarising gaze as graphs and descriptors improves accuracy over learned representations across medical tasks and GazeCopilot, reduces FHE computation by approximately $3$--$40\times$ and MPC online communication by approximately $8\times$, while facilitating interpretability.}

\subsubsection{\textbf{Interpretability of our representation}}
\label{sec:descriptor-ablation}

To assess the interpretability of our representations and the contribution of individual components, we perform a descriptor-level analysis, reporting the AUC of the full model with each descriptor removed (leave-one-out) and the AUC obtained using each descriptor independently. We further investigate the effect of spatial granularity by comparing gaze representations constructed using \(3\times3\) and \(10\times10\) grids.
\begin{table}[t]
\centering
\caption{Dropping a descriptor family: $\Delta$AUC $\times 10^{3}$ relative to the full model (negative = the family helps).}
\label{tab:family-ablation}
\begin{tabular}{lrrrr}
\toprule
\textbf{Family dropped} & \textbf{Medical} & \textbf{EMIP} & \textbf{GC read.} & \textbf{GC compr.} \\
\midrule
Size (coverage, edges, density)
& $-0.3$ & \cellcolor{green!20}$+6.0$ & $-0.1$ & \cellcolor{green!20}$+1.0$ \\

Staying/returning (self-loops, revisits)
& \cellcolor{green!20}$+0.3$ & \cellcolor{red!20}$-5.8$ & $0.0$ & \cellcolor{red!20}$-4.2$ \\

Hubs (degrees, fan-in/out)
& $+0.1$ & \cellcolor{green!20}$+1.0$ & $-0.4$ & \cellcolor{green!20}$+0.7$ \\

Loops (reciprocity, cycles, SCC)
& \cellcolor{green!20}$+0.3$ & \cellcolor{green!20}$+2.7$ & $0.0$ & \cellcolor{red!20}$-1.3$ \\

Predictability (entropies)
& \cellcolor{green!20}$+0.9$ & \cellcolor{green!20}$+0.5$ & \cellcolor{green!20}$+0.1$ & \cellcolor{green!20}$+2.9$ \\

Text Reading (code)
& -- & \cellcolor{green!20}$+6.0$ & \cellcolor{red!20}$-132.2$ & \cellcolor{green!20}$+1.1$ \\
\bottomrule
\end{tabular}
\end{table}
\begin{table}[t]
\centering
\caption{Grid resolution of the gaze-transition graph (mean AUC per domain).}
\label{tab:grid-ablation}
\begin{tabular}{lccccc}
\toprule
\textbf{Domain} &  \textbf{$3\times3$} & \textbf{$4\times4$} & \textbf{$5\times5$} & \textbf{$7\times7$} & \textbf{$10\times10$} \\
\midrule
Medical      & .738 & .740 & .739 & .740 & .740 \\
Programming  & .605 & .609 & .610 & .612 & .605 \\
\bottomrule
\end{tabular}
\end{table}
\textbf{No single descriptor matters on its own.} 
Removing any one of the graph descriptors changes the mean medical AUC by at most $\pm0.001$, and the programming mean by at most $\pm0.003$. 
On their own, graph descriptors reach 0.50--0.56 AUC in the medical tasks; the strongest are transition entropy (0.558), mean degree (0.557) and density (0.557). 
The largest single-task value is the edge count on EMS (0.659). 
Among the code-reading descriptors, the vertical reading position (0.559) and line coverage (0.558) are the strongest alone. 
Line coverage is also the only descriptor whose removal costs more than 0.004 AUC on average ($-0.008$), driven entirely by GazeCopilot readability. 
Such small, diffuse effects mean the descriptors encode overlapping aspects of the same scanpath, so pruning them one at a time is safe and attributions should be read at the family level.

With gaze statistics present, the \textbf{graph-cell resolution does not matter}: every grid from $3\times3$ to $10\times10$ yields 0.738--0.740 (medical) and 0.605--0.612 (programming; Table~\ref{tab:grid-ablation}). 
Finer grids would add no accuracy here and would only make the plaintext graph construction more expensive.

Thus, \textbf{our descriptors provide informative signals for task-specific optimization. Nevertheless, less predictive descriptors can remain valuable when more informative ones are unavailable, reflecting the heterogeneous and complementary nature of the data across tasks.}

\subsubsection{\textbf{Task-adaptive vs. per-task models}}\label{sec:ev-pred}
We further evaluate each CoGITAtion instantiation against a per-task counterpart with the same model family to evaluate the effect of deploying one model vs. multiple models for each task: CoGITAtion-NN (one task-adaptive network for all tasks) vs. a per-task NN with the same encoder, trunk and head architecture trained on one task alone;
and CoGITAtion-Tree (shared trees plus task trees with folded public gates) vs. a per-task XGBoost with the same depth-3 trees (Tables~\ref{tab:ev-gin-acc}).

CoGITAtion-Tree matches per-task tree in accuracy (mean AUC 0.754 vs. 0.755 on the medical tasks and 0.615 vs. 0.618 on programming) but evaluates twice as many trees per task: up to 50 folded shared trees plus 50 task trees, against 50 trees. 
Its circuit therefore needs about twice the bootstraps under FHE (1{,}348--1{,}467 vs. 638--743 PBS) and twice the comparisons, multiplications and communication under MPC (21{,}168 vs.\ 10{,}429 multiplications; 1{,}390 vs.\ 685\,kB per party). 
Because all comparisons of an ensemble are batched, both need the same 15 rounds, so on a WAN, where latency is dominated by rounds, the overhead almost disappears (0.86 vs.\ 0.81\,s). 
In pools whose tasks hold different owner blocks, public gating removes part of the extra trees at no accuracy cost.
What CoGITAtion-Tree buys for this overhead is deployment: one model, one compilation and one key set for all tasks and owners, availability gates that handle absent owners, and knowledge shared across tasks.

CoGITAtion-NN is \emph{cheaper} than the per-task NN. 
Skipping blocks removes encoder activations, so a pruned route needs 84--121 PBS against 121--201. 
Masking features removes secret-weight products under MPC, so it needs 42\% fewer multiplications (13{,}114 vs.\ 22{,}516) and 42\% less communication (825 vs.\ 1{,}417\,kB per party). 
Both networks need the same 41 sequential rounds, so their WAN latency is similar (2.12 vs.\
2.17\,s). 
The price is accuracy: importance estimated on small validation sets sometimes drops informative blocks, and the pruned routes fall below the per-task MLP (0.651 vs.\ 0.678 on the medical tasks). 
Without pruning the shared network is more accurate than the per-task MLP where data are scarce (0.731 vs.\ 0.699 on the medical pool; 0.722 vs.\ 0.580 on ADHD) at the same cost.

Overall, \textbf{task-adaptive CoGITAtion is the choice when one deployable model must serve heterogeneous tasks and owners, with on-par accuracy with the per-task counterpart.}

\subsubsection{\textbf{Collaboration and Vertical Data Distribution}}

We next evaluate our primary motivation: collaborative inference over vertically distributed data. 
To test collaboration on every dataset, we split each person's features vertically across three data owners.
We compare each owner alone with two ways of combining them: \emph{late fusion}, which averages the three owners' predictions and needs no computation across owners, and CoGITAtion (\emph{early fusion}), the collaborative model that takes all owners' blocks as input. 
\begin{table}[t]
\centering
\caption{Per-owner vs. collaborative performance (sample AUC).
Late: mean of the three owners' predictions; Early: one
collaborative model on all owners' blocks.}
\label{tab:ev-owners3}
\begin{tabular}{@{}lccccc@{\hspace{8pt}}|lccccc@{}}
\toprule
& \multicolumn{5}{c}{\textbf{Medical}} &
& \multicolumn{5}{c}{\textbf{Programming and GazeCopilot}} \\
\cmidrule(lr){2-6}\cmidrule(lr){8-12}
\textbf{Task} & \textbf{A} & \textbf{B} & \textbf{C} & \textbf{Late} & \textbf{Early} &
\textbf{Task} & \textbf{A} & \textbf{B} & \textbf{C} & \textbf{Late} & \textbf{Early} \\
\midrule
Duan (ASD)
& 0.703 & 0.675 & 0.619 & 0.731 & \textbf{0.741}
& EMIP exp. (rect.)
& 0.511 & \textbf{0.594} & 0.521 & 0.564 & 0.565 \\

Cilia (ASD)
& 0.704 & 0.664 & 0.598 & 0.724 & \textbf{0.730}
& EMIP exp. (vehicle)
& 0.484 & \textbf{0.597} & 0.589 & 0.587 & 0.583 \\

EMS
& 0.810 & 0.688 & 0.670 & 0.791 & \textbf{0.815}
& EMIP corr. (rect.)
& 0.439 & \textbf{0.524} & 0.517 & 0.494 & 0.486 \\

ADHD off-med.
& 0.681 & 0.786 & 0.517 & 0.759 & \textbf{0.797}
& EMIP corr. (vehicle)
& 0.460 & \textbf{0.503} & 0.457 & 0.458 & 0.444 \\

ADHD on-med.
& 0.590 & 0.747 & 0.539 & 0.739 & \textbf{0.779}
& GC readability
& 0.560 & 0.821 & 0.953 & 0.927 & \textbf{0.954} \\
&&&&&
& GC comprehension
& 0.592 & 0.628 & 0.621 & 0.670 & \textbf{0.698} \\
\bottomrule
\end{tabular}
\end{table}
\textbf{The collaborative model beats every single owner}, except for the EMIP tasks, owner B alone outperforms the collaborative model. 
Through further investigation, we found out that its block mainly contains the gaze spatial data, such as mean gaze position, in addition to the noisy nature of the dataset that we identified above.

Early fusion costs about the same as a single owner under FHE (624--743 vs.\ 600--743 PBS), because a tree ensemble has a similar number of nodes whatever its inputs.
Collaboration only adds encrypted inputs. 
Late fusion needs one circuit per owner, three times the work (1{,}852--2{,}224 PBS), but no computation across owners: each owner's circuit can run under its own key, and only the three encrypted scores are combined. 
Early fusion requires all owners' blocks in one computation, i.e.\ a common key (FHE) or joint secret sharing (MPC). Thus, \textbf{early fusion (ours) is generally more computationally efficient with better task performance.}

\subsubsection{\textbf{External Benchmark}}
\label{sec:ablation_external}

Table~\ref{tab:ev-published} compares CoGITAtion with published results on the same datasets, when available.
\begin{table*}[t]
\centering
\caption{Our models vs.\ published results on the same datasets.}
\label{tab:ev-published}
\begin{tabular}{@{}lllcc@{}}
\toprule
\textbf{Dataset} & \textbf{Method} & \textbf{Model} & \textbf{Acc.} & \textbf{AUC} \\
\midrule
Duan (ASD)
& Liaqat et al.\ 2021~\cite{liaqat2021predicting}
& ResNet-18 & 0.661 & 0.70 \\
& Liaqat et al.\ 2021
& ResNet-50 & 0.660 & 0.71 \\
& Liaqat et al.\ 2021
& 5-layer LSTM on fixations & 0.672 & 0.73 \\
& Liaqat et al.\ 2021
& MLP & 0.667 & 0.72 \\
& \textbf{CoGITAtion-Tree (ours)}
&   & 0.685 & 0.740 \\
& \textbf{CoGITAtion-NN (ours)}
&  & 0.661 & 0.712 \\
\midrule
Cilia (ASD)
& Cilia et al.\ (as reported in~\cite{zhang2026glcf})
& CNN & 0.71 & -- \\
& Zhang et al.\ 2026~\cite{zhang2026glcf}
& GLCF-Net (ViT-B/16 + CNN) & 0.870 & 0.937 \\
& \textbf{CoGITAtion-Tree (ours)}
&   & 0.825 & 0.909 \\
& \textbf{CoGITAtion-NN (ours)}
&   & 0.643 & 0.718 \\
\midrule
EMS (SZ)
& EDB\_SVM~\cite{song2025ems}
& SVM & 0.731 & 0.809 \\
& ESR\_RF~\cite{song2025ems}
& random forest & 0.763 & 0.852 \\
& DVP~\cite{song2025ems}
& DNN & 0.788 & 0.852 \\
& LVA~\cite{song2025ems}
& DNN & 0.819 & 0.899 \\
& MSNet~\cite{song2025ems}
& DNN & 0.831 & 0.897 \\
& \textbf{CoGITAtion-Tree (ours)}
&   & 0.763 & 0.877 \\
& \textbf{CoGITAtion-NN (ours)}
&   & 0.756 & 0.825 \\
\bottomrule
\end{tabular}
\end{table*}
Our models have about $10^3$ parameters: 680--730 for a per-task ensemble, 1{,}250--1{,}450 for
a CoGITAtion-Tree circuit and 1{,}240--1{,}340 for a CoGITAtion-NN route. 
The published deep models are two to five orders of magnitude larger, from about 0.12\,M parameters (branched MLP) to 11.7--86\,M (ResNet and ViT backbones).
They also operate on images or raw sequences, which makes encrypted inference impractical, whereas CoGITAtion evaluates a few hundred encrypted comparisons on descriptors computed locally by each owner. 
The only exception is SVM for ADHD which is also small; its 783 time-series features can be computed in plaintext, as ours are, but CoGITAtion still performs better. 
\textbf{CoGITAtion thus reaches accuracy comparable to published methods with a model small enough to run under FHE and MPC.}
We refer the reader to further ablation studies in Section \ref{appendix}

\section{Discussion}\label{sec:discussion}
The results show that \textbf{CoGITAtion can jointly address predictive performance, interpretability, and privacy-preserving computation efficiency for heterogeneous gaze data.} 
By converting variable-length gaze recordings into compact, interpretable descriptors, the framework substantially reduces the complexity of private inference while retaining competitive predictive performance. 

\paragraph{\textbf{Limitations and Future Work.}}
While the gaze-transition descriptors provide a compact representation well suited to FHE and MPC, summarising a scanpath inevitably removes some temporal information. 
This is reflected in the EMIP results, where learned features outperform CoGITAtion, suggesting that ordered fixation information remains important for some code-reading tasks \cite{bednarik2020emip}. 
Future work could therefore incorporate lightweight temporal descriptors while preserving the compact structure required for private inference. 
\\
The collaboration experiments support the value of combining complementary owner-specific information. 
However, several limitations remain. 
The vertical data-ownership experiments simulate distributed ownership by partitioning existing feature blocks rather than using naturally vertically distributed institutional datasets. 
Hence, future work should focus on developing vertically distributed gaze datasets to better evaluate privacy-preserving collaborative learning across data owners with heterogeneous feature spaces.
\\
Overall, our results suggest that \textbf{representation design is central to practical privacy-preserving gaze analysis}.

\section{Conclusion}\label{sec:conclusion}
This work presents CoGITAtion, a framework for privacy-preserving gaze analysis that combines compact, interpretable gaze-transition descriptors with task-adaptive architectures and secure inference protocols. 
By transforming variable-length gaze recordings into fixed-dimensional representations locally at each data owner, CoGITAtion reduces the computational burden of encrypted inference while preserving predictive information across heterogeneous tasks. 
It further supports model reuse and accommodates heterogeneous or missing feature blocks through availability-aware routing and feature selection, avoiding unnecessary computation during private inference. 
Overall, CoGITAtion demonstrates how compact representations, adaptive computation, and cryptographic protection can be jointly leveraged to make privacy-preserving gaze analysis more efficient, interpretable, and practical across diverse applications, while providing a framework to "cogitate" the joint design of these components and systematically explore their synergies and trade-offs.


\section*{Ethics and Privacy Statement}
All datasets used in this study are publicly available. 
CoGITAtion is designed primarily to enable privacy-preserving gaze analysis by protecting sensitive gaze data during collaborative inference.

\bibliographystyle{ACM-Reference-Format}
\bibliography{sample-base}

\newpage
\appendix
\section{Ablation Studies}\label{appendix}

We conduct a comprehensive set of ablation studies to quantify the contribution of the main design choices in CoGITAtion. 
The ablations examine six aspects of the framework: 
(i) the shared--task-specific decomposition and public gating in CoGITAtion-Tree, 
(ii) task adaptation and feature pruning in CoGITAtion-NN, 
(iii) the effect of pooling heterogeneous tasks into a single model, 
(iv) the contribution of individual owner feature blocks, 
(v) potential stimulus confounds, and 
(vi) the benefit of collaboration between vertically distributed data owners. 

\subsection{Shared and Task-Specific Trees and Gating}
\label{sec:ablation_tree}

The first ablation evaluates two central design choices of CoGITAtion-Tree:
(1) whether a multi-task model should use a single pooled ensemble or explicitly separate shared and task-specific components, and
(2) whether publicly known task and owner information can be exploited to reduce encrypted inference cost without changing the model output.

We compare three tree-based configurations. 
The first is a single 50-tree ensemble of depth three that receives the task identity as an input together with the available feature blocks. 
The second increases the capacity of this pooled model to 150 trees with depth five, testing whether a larger single ensemble can compensate for the lack of explicit task specialisation. 
The third is the proposed CoGITAtion-Tree architecture, which contains 50 shared trees trained without the task indicator and 50 task-specific trees for each task. 
The shared trees therefore model information common across tasks, while the task-specific trees capture residual task-dependent structure.

For every configuration, we measure both predictive performance and the cost of encrypted inference. We report the number of programmable bootstraps (PBS) before and after public gating. 
Public gates correspond to information known before private inference, including the task identity, owner-block availability, and the values used for absent blocks. 
Splits depending only on these public values can therefore be folded at compile time.

\begin{table}[t]
\centering
\caption{Ablation of multi-task ensemble design and public gating. AUC is averaged over the tasks in each pool. PBS denotes the encrypted inference cost before and after compile-time public gating, with the percentage reduction in PBS.}
\label{tab:ablation_tree}
\begin{tabular}{lccc}
\toprule
\textbf{Task pool} &
\textbf{Single, 50 trees} &
\textbf{Single, 150 trees, depth 5} &
\textbf{CoGITAtion-Tree} \\
\midrule
ASD
& 0.719 (735 $\rightarrow$ 712; 3\%)
& 0.722 (7,289 $\rightarrow$ 7,014)
& \textbf{0.732} (1,450; 0\%) \\
ADHD
& 0.746 (735 $\rightarrow$ 730; 1\%)
& 0.740 (7,151 $\rightarrow$ 7,005)
& \textbf{0.747} (1,451; 0\%) \\
Medical
& 0.725 (742 $\rightarrow$ 518; 30\%)
& 0.743 (7,691 $\rightarrow$ 5,315)
& \textbf{0.756} (1,465 $\rightarrow$ 1,248; 15\%) \\
Programming
& 0.608 (739 $\rightarrow$ 637; 14\%)
& 0.614 (7,656 $\rightarrow$ 5,734)
& \textbf{0.615} (1,414; 0\%) \\
All 11 tasks
& 0.633 (743 $\rightarrow$ 497; 33\%)
& 0.665 (8,021 $\rightarrow$ 4,381)
& \textbf{0.677} (1,443 $\rightarrow$ 1,231; 15\%) \\
\bottomrule
\end{tabular}%
\end{table}

As shown in Table~\ref{tab:ablation_tree}, CoGITAtion-Tree achieves the highest AUC in every task pool. 
The difference is particularly apparent as the task pool becomes more heterogeneous. 
Increasing the capacity of the pooled ensemble partially closes this gap but does not eliminate it. 
This increase in capacity, however, comes at substantially higher encrypted inference cost. 
These results indicate that the benefit of CoGITAtion-Tree does not simply arise from having a larger ensemble. 
Instead, explicitly separating task-independent and task-specific structure  allows the model to allocate capacity according to the heterogeneous task requirements. 
The shared trees capture patterns that can be reused across tasks, while the task-specific trees provide additional capacity only where it is required.

Public gating provides a lossless reduction in encrypted computation. 
Across all tasks, folding the public gates produced exactly the same model output as the corresponding ungated model.
Thus, the optimisation does not approximate the original model or modify its predictions.
The magnitude of the reduction depends on the heterogeneity of the task pool. 
For CoGITAtion-Tree, gating reduces the PBS count by 15\% for the medical pool.
In contrast, no reduction is obtained for the ASD, ADHD, or programming pools because the tasks within these groups use the same feature blocks. 
Interestingly, the single pooled ensembles benefit more strongly from gating in the heterogeneous pools, with reductions of up to 30\%. 
This occurs because these models explicitly use the task indicator in their trees, resulting in branches whose sole purpose is to distinguish between tasks. 
Once the task is publicly known, such branches can be folded away. 
Despite this larger computational reduction, these models remain less accurate than CoGITAtion-Tree.

The ablation concludes that the \textbf{shared-task-specific decomposition provides the strongest predictive performance without requiring the very large ensembles needed by a single pooled model.} 
Public gating is an exact compile-time optimisation that reduces encrypted computation whenever task or owner availability is heterogeneous.
\textbf{The benefits of the gating mechanism increase with the heterogeneity of the data and tasks, yielding greater performance reductions in more heterogeneous settings.}

\subsection{Shared and Task-Specific NN and Gating}
\label{sec:ablation_nn}
We next examine the individual contributions of the adaptation mechanisms in CoGITAtion-NN.
For each task pool, we compare five variants of the same trained network. 
The \emph{full route} uses all available blocks without feature masking. 
\emph{Skip + mask} removes uninformative blocks and masks unimportant features selected using an inner validation split. 
The \emph{+adapt} variant additionally freezes the shared components and retrains the task-specific components. 
We further simulate 6-bit quantisation of the pruned route. 
Finally, the \emph{task-private} configuration evaluates the full network with all task heads, preventing the server from learning which task is being evaluated.

\begin{table}[t]
\centering
\caption{Ablation of task adaptation mechanisms in CoGITAtion-NN. 
AUC is averaged over the corresponding task pool. 
PBS and encrypted-input counts are reported before and after pruning where applicable.}
\label{tab:ablation_nn}
\begin{tabular}{lccccc}
\toprule
\textbf{Pool} &
\textbf{Full} &
\textbf{Skip+mask} &
\textbf{+adapt} &
\textbf{6-bit} &
\textbf{PBS full $\rightarrow$ pruned} \\
\midrule
ASD
& 0.702 & 0.648 & 0.656 & 0.654 & 121 $\rightarrow$ 96 \\
ADHD
& 0.722 & 0.563 & 0.582 & 0.575 & 201 $\rightarrow$ 98 \\
Medical
& 0.731 & 0.651 & 0.658 & 0.656 & 161 $\rightarrow$ 103 \\
Programming
& 0.624 & 0.594 & 0.597 & 0.596 & 201 $\rightarrow$ 101 \\
\bottomrule
\end{tabular}%
\end{table}

\begin{table}[t]
\centering
\caption{Encrypted input and task-private inference costs for the CoGITAtion-NN ablation.}
\label{tab:ablation_nn_cost}
\begin{tabular}{lccc}
\toprule
\textbf{Pool} &
\textbf{Encrypted inputs} &
\textbf{Task-private PBS} &
\textbf{PBS after pruning} \\
\midrule
ASD
& 37 $\rightarrow$ 13 & 162 & 96 \\
ADHD
& 59 $\rightarrow$ 8.4 & 242 & 98 \\
Medical
& 47 $\rightarrow$ 10.3 & 365 & 103 \\
Programming
& 56 $\rightarrow$ 8.7 & 406 & 101 \\
\bottomrule
\end{tabular}
\end{table}

Skipping and masking substantially reduce encrypted computation. 
PBS is approximately halved for the ADHD and programming pools, while the number of encrypted inputs decreases by approximately 3--7$\times$. 
The two mechanisms affect different components of the encrypted computation: block skipping removes the corresponding encoder activations and therefore reduces PBS, whereas feature masking primarily reduces the number of encrypted inputs.
Howver, the computational savings come at an accuracy cost. 
For example, the largest degradation occurs for ADHD, where AUC decreases from 0.722 to 0.563 after skipping and masking. 
The degradation is smaller after task-specific adaptation, which recovers between 0.007 and 0.019 AUC across the evaluated pools. 
Six-bit quantisation introduces little additional degradation, with a maximum AUC change of 0.007.
\textbf{The task-private configuration increases the computational cost substantially, requiring 162--406 PBS compared with 121--201 PBS for a single full route. 
This reflects the cost of evaluating all task heads when the task identity itself must remain private.}

These results establish task adaptation as a computational cost-accuracy trade-off rather than an unconditional improvement in predictive performance. 
On the relatively small cohorts considered here, importance estimates obtained from a limited number of validation participants can be unstable, causing informative blocks or features to be removed. 
Consequently, the full route provides the most reliable accuracy, whereas pruning is most useful when encrypted-input or computational constraints are dominant.

\end{document}